\documentclass[]{ceurart}

\usepackage{listings}
\usepackage{todonotes}
\usepackage{tikz}
\usepackage{enumitem}
\usepackage{hyperref}
\usetikzlibrary{shapes.geometric, arrows}

\usepackage{xcolor}

\lstdefinestyle{codeblock}{
    basicstyle=\ttfamily\small,
    keywordstyle=\color{blue!70!black},
    commentstyle=\color{gray!80},
    stringstyle=\color{green!50!black},
    showstringspaces=false,
    breaklines=true,
    frame=single,
    framerule=0.2pt,
    backgroundcolor=\color{gray!5}
}

\usepackage{todonotes}

\begin{document}

\copyrightyear{2025}
\copyrightclause{Copyright for this paper by its authors.
  Use permitted under Creative Commons License Attribution 4.0
  International (CC BY 4.0).}

\conference{QKG @ ESWC'26}

\title{A Benchmark for Gap and Overlap Analysis as a Test of KG Task Readiness}
\author[1]{Maruf Ahmed Mridul}[
email=mridum@rpi.edu,
orcid=0009-0003-7501-4714
]

\author[2]{Rohit Kapa}[
email=rohit.kapa@prudential.com
]

\author[1]{Oshani Seneviratne}[
email=senevo@rpi.edu,
orcid=0000-0001-8518-917X
]

\address[1]{Rensselaer Polytechnic Institute, Troy, New York, United States}
\address[2]{Prudential Financial, Inc.}

\begin{abstract}
  Task-oriented evaluation of knowledge graph (KG) quality increasingly asks whether an ontology-based representation can answer the competency questions that users actually care about, in a manner that is reproducible, explainable, and traceable to evidence. This paper adopts that perspective and focuses on gap and overlap analysis for policy-like documents (e.g., insurance contracts), where given a scenario, which documents support it (overlap) and which do not (gap), with defensible justifications. The resulting gap/overlap determinations are typically driven by genuine differences in coverage and restrictions rather than missing data, making the task a direct test of KG task readiness rather than a test of missing facts or query expressiveness. We present an executable and auditable benchmark that aligns natural-language contract text with a formal ontology and evidence-linked ground truth, enabling systematic comparison of methods. The benchmark includes: (i) ten simplified yet diverse life-insurance contracts reviewed by a domain expert, (ii) a domain ontology (TBox) with an instantiated knowledge base (ABox) populated from contract facts, and (iii) 58 structured scenarios paired with SPARQL queries with contract-level outcomes and clause-level excerpts that justify each label. Using this resource, we compare a text-only LLM baseline that infers outcomes directly from contract text against an ontology-driven pipeline that answers the same scenarios over the instantiated KG, demonstrating that explicit modeling improves consistency and diagnosis for gap/overlap analyses. Although demonstrated for gap and overlap analysis, the benchmark is intended as a reusable template for evaluating KG quality and supporting downstream work such as ontology learning, KG population, and evidence-grounded question answering. The benchmark and accompanying codebase are available at \url{https://github.com/brains-group/gap_and_overlap_analysis_insurance_contract_benchmark}.
\end{abstract}


\maketitle

\section{Introduction}

Assessing and improving the quality of Knowledge Graphs (KGs) is an active research area, motivated by the growing use of KGs as reusable and interoperable resources for downstream analysis and AI systems ~\cite{xue2022knowledge,tsaneva2025knowledge}. In addition to structural and logical criteria, many evaluation efforts now take a \emph{task-oriented} view of quality: a representation is valuable if it supports the competency questions ~\cite{bezerra2013evaluating, araujo2016data} that users care about, with answers that are reproducible, explainable, and traceable to evidence. This paper adopts that perspective and studies a concrete task family centered on \emph{gap and overlap analysis}.

In many domains, important rules and exceptions are written in \emph{policy-like documents}, such as contracts, regulations, service terms, or product specifications. These documents state what is supported under certain conditions, what is excluded or denied, and what is outside scope. Stakeholders frequently need to compare multiple documents: for a given scenario, which documents support it (overlap), and which do not (gap). For example, a scenario in which an insured dies by suicide 13 months after policy issuance would be covered by contracts with a 12-month exclusion period but not by those with a 24-month exclusion. The overlapping contracts are those that agree on covering it, while the gap is the subset that do not. Here, gaps and overlaps are usually \emph{substantive}: they occur because documents differ in their intended coverage and restrictions, not because a KG is missing facts or because a query cannot be written. From a KG perspective, gap and overlap analysis is therefore a useful test of task readiness: can an ontology-based representation show these differences consistently, and can it provide evidence that supports each outcome?

Reproducible evaluation of such analyses is nevertheless difficult without aligned resources that connect natural language documents to formal representations and queryable ground truth. In practice, it is common for such document collections to be complex, hard to share, or insufficiently annotated for detailed benchmarking, which limits the ability to evaluate both explainability and correctness of the results produced by different methods. It also limits the evaluation of automatic ontology generation and refinement methods, since their outputs cannot be tested against shared, executable competency questions with evidence. To address this barrier, we contribute a benchmark dataset that supports such evaluation purposes, and we demonstrate its use through a workflow that makes gap and overlap analysis executable and auditable. In this paper, we focus on the life insurance domain, where contracts are legally structured yet heterogeneous in their benefits, exclusions, conditions, and temporal constraints, making them a suitable setting for auditable gap and overlap analysis.

Our contributions are four-fold. 
\begin{itemize}
    \item First, we release a manually curated corpus of ten simplified but diverse \emph{life insurance contracts}, verified by a domain expert (who is also a co-author on this paper), designed to cover a range of clause patterns while remaining manageable for systematic study.
    \item Second, we provide a domain ontology (TBox) that formalizes the main contract concepts and relations, together with an aligned ABox populated from the contract facts, enabling deterministic, query-based analysis.
    \item Third, we release a suite of 58 structured scenarios and a corresponding ground truth for gap and overlap analysis, which is also reviewed and verified by a domain expert. For each scenario, we include contract-level outcomes and clause-level source excerpts that justify those outcomes, supporting traceability and diagnosis.
    \item Fourth, we evaluate gap and overlap determination using two approaches: a text-only LLM pipeline that infers contract responses for each scenario, and an ontology-driven pipeline that answers the same scenarios using SPARQL over the instantiated ontology. This comparison measures the value of explicit ontological modeling for producing consistent and auditable gap and overlap analyses.
\end{itemize}
   
Although the benchmark is built in the life insurance domain, the main goal is more general. It shows how aligned document text, an ontology with instances, scenario probes, and evidence-linked labels can form a reusable resource for KG quality evaluation and for downstream tasks such as ontology learning, KG population, and explainable question answering. More broadly, this benchmark aims to address key challenges in modern AI: the limits of LLM reasoning, the need for explainable and auditable outputs, and the integration of structured-logic based and neural methods. By providing a dataset where every answer is traceable to structured evidence, we enable faithful evaluation of reasoning systems beyond surface-level accuracy. Our results demonstrate that while LLMs can approximate coverage judgments, they are less reliable when a scenario requires strict handling of structural applicability, contractual scope, or clause-level justification. By contrast, the ontology-backed pipeline makes these commitments explicit and executable. In this sense, the benchmark is not only a resource for KG quality evaluation, but also a controlled testbed for studying how structured, logic-based representations can complement neural methods in high-stakes domains where correctness and justification matter.

\section{Related Work}

KG quality evaluation has received attention since Zaveri et al.~\cite{zaveri2016quality}
systematized 18 quality dimensions for linked data, and Xue and Zou~\cite{xue2022knowledge}
extended this framework across the full KG lifecycle. Paulheim~\cite{paulheim2016knowledge}
surveys refinement approaches for correcting ABox-level errors through statistical and
embedding-based methods. More recently, Tsaneva et al.~\cite{tsaneva2025knowledge} combine
LLM suggestions with human-in-the-loop review to detect structural issues in populated KGs.
Complementing these structural and statistical approaches, recent work has explored
explainability-driven methods for improving rule-based reasoning systems.
Seneviratne et al.~\cite{seneviratne2025explainability} introduce a framework that leverages
trace-based, contextual, contrastive, and counterfactual explanations to debug and refine
rules, enabling human-in-the-loop validation of reasoning outcomes. This line of work
highlights the importance of interpretability not only for model transparency but also for
systematic quality improvement in knowledge-driven systems.

A consistent observation across this work is that structural metrics do not assess whether a
KG supports the analytical tasks it was built for. The task-oriented view of quality: a
representation is valuable if it correctly answers the competency questions that motivated its
construction~\cite{bezerra2013evaluating, araujo2016data}, follows directly from the
methodology introduced by Grüninger and Fox~\cite{gruninger1995methodology}, who argued
that an ontology's correctness should be characterized by the queries it can answer. Our
benchmark operationalizes this view: the 58 scenarios function as executable competency
questions whose outcomes serve as a task-oriented quality test for both the TBox
and ABox. Unlike structural validators such as SHACL~\cite{knublauch2017shacl}, this benchmark tests whether a KG can produce consistent and evidence-grounded coverage determinations across heterogeneous contracts. Performance directly reflects the ability to support reproducible gap and overlap determinations, with strong results implying task-scoped consistency within the defined scenario suite, though broader completeness beyond that is not guaranteed. Therefore our benchmark can be positioned as a complementary, task-oriented layer in any KG quality workflow.

The closest work to our gap and overlap task comes from the legal NLP community.
Koreeda and Manning~\cite{koreeda2021contractnli} introduce ContractNLI, a document-level
natural language inference dataset of annotated contracts, where a system must determine whether a hypothesis
is entailed by, contradicts, or is not mentioned in a contract, with supporting evidence
spans. Hendrycks et al.~\cite{hendrycks2021cuad} introduce CUAD, a large expert-annotated
dataset for identifying 41 types of salient clauses across 500+ contracts. Both works address
the problem of extracting and verifying claims against contract text, which motivates our
use case. However, neither provides cross-contract comparison with formal semantics: they
treat each contract independently and rely on neural models rather than a shared formal
representation. Kang et al.~\cite{kang2024using} similarly find that LLM-generated structured outputs from health insurance policies require human oversight, further motivating formal representations. 
This limitation has motivated prior efforts to encode contractual logic in structured,
machine-executable forms. Seneviratne et al.~\cite{seneviratne2022towards} advocate for
computable contracts in insurance marketplaces, emphasizing transparency, automation, and
auditable decision-making. Building on this vision, Van Woensel et al.~\cite{vanwoensel2023enabling}
demonstrate how clinical decision logic represented in KGs can be translated
into executable smart contracts, enabling consistent enforcement across distributed systems.
Our work is complementary: we contribute a structured representation in which the
same query evaluates all ten contracts under identical logic, producing coverage
determinations with traceable, clause-level evidence rather than span-level predictions.

The domain representation underlying this work is informed by prior efforts
to formalize financial and legal knowledge. FIBO~\cite{bennett2013financial} provides a
structured vocabulary for financial instruments, entities, and contractual
obligations, and LKIF~\cite{hoekstra2007lkif} models the structure of legislation
and regulatory rules. Both are oriented toward standardization and
interoperability at the schema level. 
Recent work has further explored how KGs can act as an intermediate
semantic layer for executable systems. Van Woensel and Seneviratne~\cite{vanwoensel2025semantic}
propose generating blockchain smart contracts directly from KGs, enabling
semantic interoperability across distributed healthcare systems while preserving alignment
with domain standards.
Our TBox draws on the same ontology
engineering principles but is purpose-built for the life insurance contract
corpus, prioritizing instance-level population and query-based evaluation. Pan et al.~\cite{pan2024unifying} frame the complementarity of KGs
and LLMs as a general research program, arguing that structured representations provide
verifiability that LLMs alone cannot guarantee. Our comparison of SPARQL-based and
text-only LLM pipelines is a concrete empirical instance of that argument in
the insurance domain.

Automated construction of ontologies and knowledge bases from text is a
rapidly growing area of research. Babaei Giglou et al.~\cite{babaei2023llms4ol}
introduce the LLMs4OL paradigm and evaluate nine LLM families on term typing,
taxonomy discovery, and non-taxonomic relation extraction, finding that while
LLMs capture ontological structure from text, performance varies substantially
by domain and task. Saeedizade and Blomqvist~\cite{saeedizade2024navigating}
show that GPT-4 can produce OWL modeling suggestions of reasonable quality
when guided by competency questions, though complex axiom patterns remain
difficult. Mihindukulasooriya et al.~\cite{mihindukulasooriya2023text2kgbench}
provide Text2KGBench, a benchmark for evaluating LLM-driven KG generation
from text, measuring whether extracted triples conform to a given schema.
Collectively, this work demonstrates that automated construction is feasible
but that evaluation remains a bottleneck: resources that provide aligned domain-specific NL documents, a verified TBox, and a populated ABox together as a gold standard against which automated pipelines can be measured are very scarce.
Our benchmark provides exactly that, and the executable scenarios with
clause-level ground truth add a further layer of functional evaluation on top
of it.

\section{The Dataset: A Benchmark for Insurance KGs}

The proposed benchmark is composed of three primary, interlinked components: (i) a curated corpus of diverse life insurance contracts, (ii) a verified domain ontology (TBox) with a corresponding instantiated knowledge base (ABox), and (iii) a comprehensive suite of executable competency questions. Together, these artifacts provide the necessary substrate for evaluating the reproducibility, explainability, and evidence-linked traceability of coverage determinations across heterogeneous contract types.

\subsection{The Contracts}

\begin{figure}[t]
    \centering
    \includegraphics[width=0.9\linewidth]{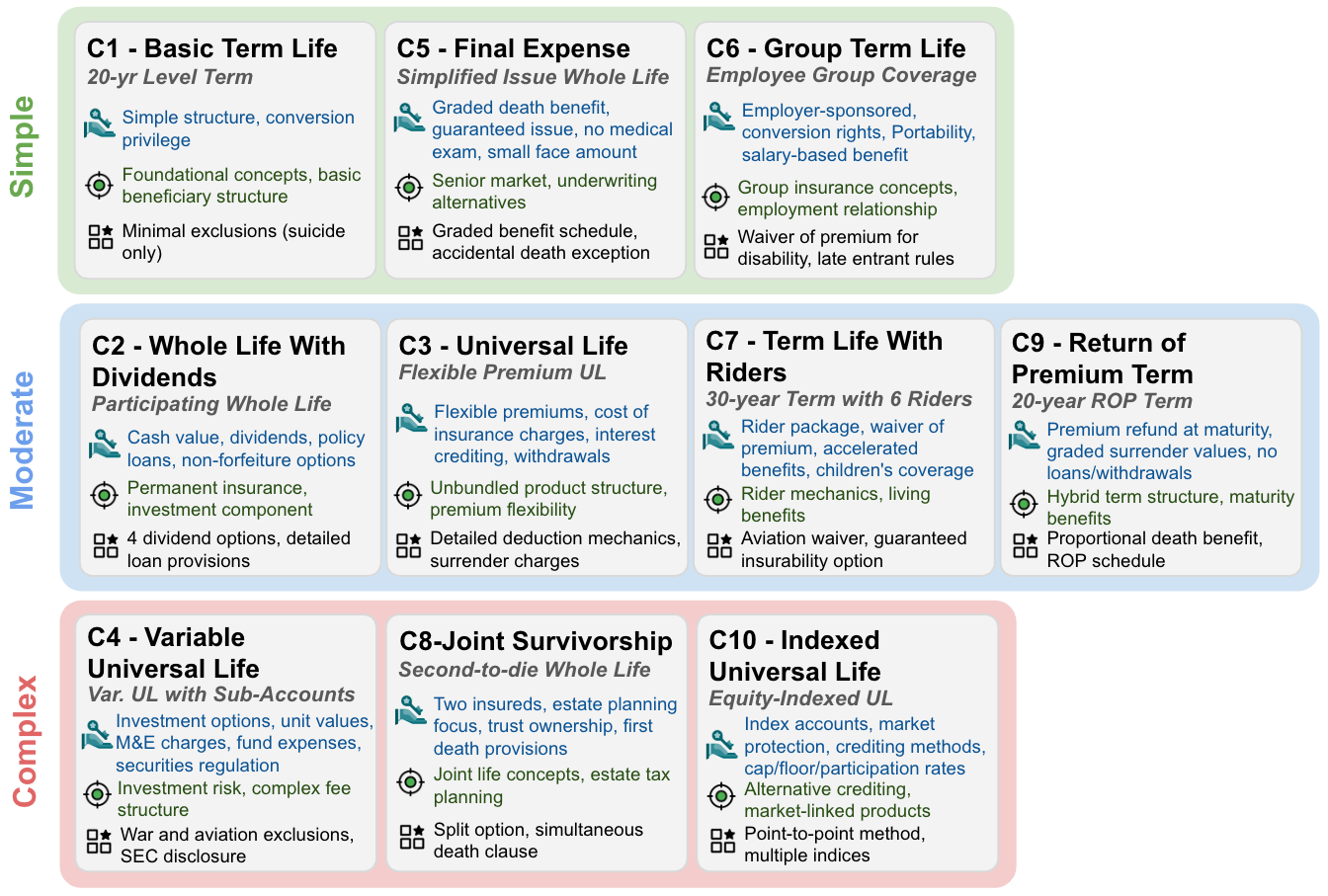}
    \caption{\textbf{Overview of the 10 life insurance contracts categorized by complexity}. The contracts are grouped into three levels: \textit{Simple}, \textit{Moderate}, and \textit{Complex}, based on their complexity and range of features. \raisebox{-0.6ex}{\includegraphics[width=0.03\textwidth]{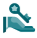}}, \raisebox{-0.6ex}{\includegraphics[width=0.032\textwidth]{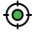}}, and \raisebox{-0.4ex}{\includegraphics[width=0.025\textwidth]{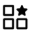}} represent the \textit{Key Features}, \textit{Focus}, and \textit{Uniqueness}, respectively. C1-C10 are the identifiers of the contracts.}
    \label{fig:contract_summary}
\end{figure}

We release ten synthetic life insurance contracts, each representing a distinct product type found in the life insurance market. The contracts were developed by first cataloging the major product categories through domain research and then generating representative contract text for each using \textit{Claude Sonnet 4.6}~\cite{anthropic2026claude46} with carefully structured prompts. Each generated contract was subsequently reviewed and validated by a domain expert to ensure terminological consistency, legal plausibility, and internal coherence across contract clauses.
Together, the ten contracts cover a wide range of product types, spanning most major categories of individual and group life insurance commonly found in the market. They range from simple term products to highly structured products such as variable and indexed universal life, providing a diverse but tractable corpus for systematic study. See Figure \ref{fig:contract_summary} for a detailed overview of each contract, highlighting their coverage, diversity, and complexity.

The set of contracts is designed with two complementary goals: breadth of coverage and tractability for systematic study. For breadth, the contracts collectively instantiate a wide variety of product structures, underwriting approaches, and provision types. For tractability, each contract is simplified enough that every substantive provision can be mapped to ontology concepts, enabling complete manual annotation. Diversity is evident not only in product type but also in structural variation across shared concepts, which is precisely the kind of variation that motivates gap and overlap analysis. For example, variations like the suicide exclusion clause: most contracts use a 24-month exclusion, but one uses 12 months, with varying benefits. Similarly, the grace period varies between premium non-payment (30-31 days) and account value depletion (61 days). These differences reveal the gaps and overlaps the benchmark seeks to highlight.

\subsection{The Ontology}

The ontology was constructed through a structured, iterative process. We first used Claude Sonnet~4.6 to generate a base OWL/Turtle ontology from the ten contracts and a detailed prompt specifying the required domain coverage and design constraints. The resulting draft was then subjected to multiple rounds of LLM-assisted review: we independently queried Claude, Gemini, and ChatGPT to identify structural issues, missing classes, incorrect property domains and ranges, and violations of  design patterns, and manually resolved each reported issue. Finally, we conducted a complete manual review of the TBox, resolving any remaining inconsistencies. The outcome is a verified ontology whose structure is grounded in standard Semantic Web practice.

\subsubsection{TBox and ABox}
The TBox defines the vocabulary and structure of the life insurance domain in OWL/Turtle. It is organized around a root \texttt{Policy} class from which all product-type subclasses descend via \texttt{rdfs:subClassOf}, covering the full range of product families represented in the corpus. Alongside the policy hierarchy, the TBox covers parties
(\texttt{Insured}, \texttt{PolicyOwner}, \texttt{Beneficiary},
\texttt{Trust}, \texttt{Employer}), coverage and benefit types,
riders, premiums, grace periods, cash value and crediting mechanisms, charges, exclusions, and provisions. All classes and properties carry \texttt{rdfs:label} and \texttt{rdfs:comment} annotations; the comments record which contracts instantiate each concept and under what conditions. Following is an example that shows the high level modeling of suicide exclusion in the life insurance contracts.

\begin{lstlisting}[style=codeblock]
li:SuicideExclusion a owl:Class ; rdfs:subClassOf li:Exclusion ;
    rdfs:label "Suicide Exclusion" ;
    rdfs:comment "Present in ALL 10 contracts. Period: 12 months (C6 only); 24 months (C1,C2,C3,C4,C5,C7,C8,C9,C10). Benefit outcome varies - use li:suicideBenefitType object property with li:SuicideBenefitType individuals." .
\end{lstlisting}

The ABox instantiates one policy individual per contract (C1--C10),
each linked through object properties to supporting individuals
representing the insured, policy owner, insurer, beneficiaries, death
benefit, premium, grace period, exclusions, provisions, riders, cash
value, charges, and crediting mechanisms.
Product-specific structures such as investment sub-accounts, index accounts,
irrevocable trust ownership, etc receive dedicated individuals where
required. The following excerpt from C1's ABox shows an example of how policy individuals are instantiated.

\begin{lstlisting}[style=codeblock]
li_abox:SuicideExclusion_C1 a li:SuicideExclusion ;
    li:suicideExclusionPeriodMonths 24 ;
    li:suicideBenefitType li:ReturnPremiumsPaid ;
    li:sourceContractID "C1" ;
    li:sourceContractSection "Section 7.1" ;
    li:coverageSourceText "SUICIDE CLAUSE: If the insured commits suicide within two years from the issue date, our liability is limited to a refund of premiums paid." ; .
\end{lstlisting}

Every ABox individual is annotated with three traceability properties:
\texttt{sourceContractID}, \texttt{sourceContractSection}, and
\texttt{coverageSourceText} (verbatim or close-paraphrase of the
originating contract clause). These annotations ensure that any SPARQL query result can be traced directly to the supporting contract text, which is a prerequisite for explainable gap and overlap analysis.
Instance density varies with product complexity: simpler term policies
require fewer individuals covering their essential provisions, while the
variable universal life and indexed universal life policies require
substantially more to represent sub-accounts, multiple charge types,
exclusions, and crediting mechanisms in full.

\subsubsection{Qualitative Analysis}

The correctness of gap and overlap determinations in this context depends heavily on the precision of the ontology's structural commitments. Poorly defined class boundaries or inconsistent property representations can cause contracts with genuinely different clauses to produce identical query results, obscuring real gaps or introducing false overlaps. A well-structured TBox ensures that parametric differences such as exclusion periods or benefit types are encoded as typed, queryable values, enabling deterministic, evidence-linked analysis. Therefore, our TBox design choices follow established  ontology engineering practices while
introducing domain-specific qualities that make it suitable as a ground-truth reference. Its class hierarchy is organized around the \textbf{subsumption pattern}, with
\texttt{Policy}, \texttt{Party}, \texttt{Coverage}, \texttt{Exclusion},
\texttt{Provision}, and \texttt{Rider} serving as abstract superclasses from which domain-specific subclasses inherit. Depth is proportional to domain complexity: the \texttt{Policy} branch reaches
four levels (e.g., \textit{Policy\;$\to$\;PermanentLifePolicy\;$\to$\;UniversalLifePolicy\;$\to$\;IndexedUniversalLifePolicy}), while shallower branches reflect concepts where the contracts do not require further distinction.
Party roles such as \texttt{FirstInsured}, \texttt{SecondInsured},
\texttt{IrrevocableTrust}, \texttt{RevocableTrust}, etc., are modeled as distinct
classes rather than collapsed into generic role-bearing individuals, keeping
property domain and range constraints precise.

Where a concept admits a fixed set of mutually exclusive values, the TBox
applies the \textbf{value partition pattern}: named individuals typed against a
dedicated class replace free-text literals. For instance, the \texttt{SuicideBenefitType} class enumerates five named individuals, each
mapped to specific contracts via a typed object property, enabling a reasoner or
SPARQL endpoint to test for contract behaviors formally rather than through
string matching.
This process also maintains a clean \textbf{TBox/ABox separation}: the TBox defines only
classes, properties, and controlled-vocabulary individuals, leaving all
contract-specific assertions to the ABox, which allows the schema to function as
a stable, reusable vocabulary against which multiple ABoxes can be independently
validated.

Object and datatype properties carry explicit \texttt{rdfs:domain} and \texttt{rdfs:range} declarations, with datatype properties bound to appropriate XSD types (\texttt{xsd:integer}, \texttt{xsd:decimal}, \texttt{xsd:boolean}, \texttt{xsd:date}). This type discipline is functional: properties such as
\texttt{exclusionIsTimeBounded}, \texttt{exclusionPeriodMonths}, and
\texttt{suicideExclusionPeriodMonths} work in combination to let a reasoner
distinguish permanent from time-bounded exclusions without parsing narrative
text.

At the schema level, every class and property carries an \texttt{rdfs:comment} annotation recording which contracts instantiate it and under what parametric conditions, including cross-contract variation and known modeling ambiguities, serving as human-readable documentation embedded directly in the ontology. More importantly, this annotation is operationalized at the ABox level through three dedicated datatype properties defined in the TBox: \texttt{sourceContractID}, \texttt{sourceContractSection}, and \texttt{coverageSourceText}. These are declared with \texttt{rdfs:domain owl:Thing}, making them applicable to any instance, and are required on every ABox individual to anchor each assertion to a specific contract identifier, section reference, and supporting contract language. This design ensures that every populated fact in the knowledge base is traceable to its documentary source, which is particularly valuable in legal and regulatory domains where the auditability of extracted knowledge matters.

\begin{wrapfigure}{r}{0.65\textwidth}
  \centering
  \includegraphics[width=\linewidth]{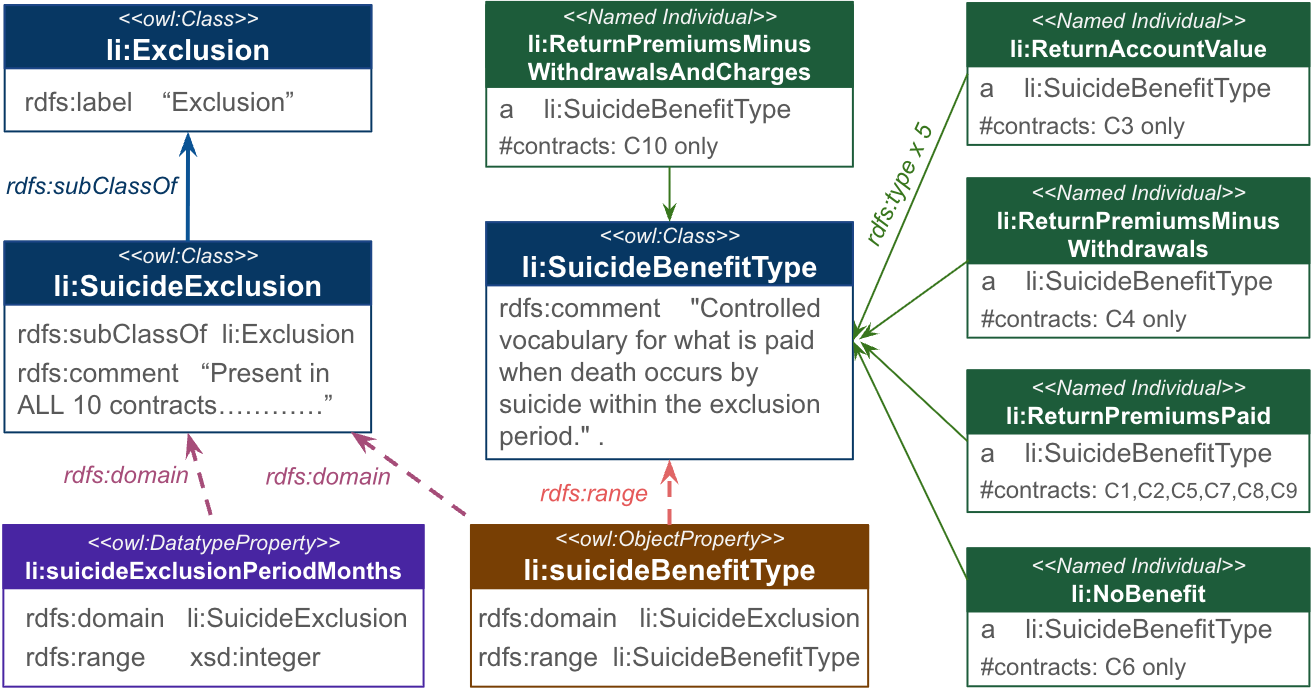}
  \caption{A representative TBox excerpt.}
  \label{fig:ontology-excerpt}
\end{wrapfigure}

Figure~\ref{fig:ontology-excerpt} presents a representative excerpt that jointly
illustrates the qualities described above. The \texttt{Exclusion}
superclass and \texttt{SuicideExclusion} subclass demonstrate the subsumption
pattern; the \texttt{SuicideBenefitType} class with its four named individuals
demonstrates the value partition pattern; the \texttt{suicideBenefitType} object
property and \texttt{suicideExclusionPeriodMonths} datatype property demonstrate
typed property constraints with explicit domain and range declarations; and the
\texttt{rdfs:comment} on \texttt{SuicideExclusion}, recording per-contract
exclusion periods across all ten contracts, demonstrates the annotation practice.

\subsubsection{Ontology Coverage Analysis}

To assess how well the TBox captures the domain vocabulary expressed in the contracts, we conducted a keyphrase-based coverage analysis. The keyphrases were extracted from each contract using Claude Sonnet~4.6, resulting in a set of 416~phrases drawn across all ten contracts.
These were matched against human-readable TBox artifacts:  classes, properties, and labels using three complementary methods:
\begin{enumerate}[nosep,leftmargin=*]
  \item \textbf{Exact / substring matching} --- all component words of a
        keyphrase are present in the target label
        (e.g., \emph{face amount} $\to$ \texttt{faceAmount});
  \item \textbf{Fuzzy matching} --- RapidFuzz\footnote{https://pypi.org/project/RapidFuzz} token-sort ratio
        $\geq 70$ (e.g., \emph{current credited interest rate} $\to$
        \texttt{creditingCurrentRate});
  \item \textbf{Semantic similarity} --- SentenceTransformer
        \texttt{all-MiniLM-L6-v2} \footnote{https://pypi.org/project/sentence-transformers} cosine similarity $\geq 0.70$
        (e.g., \emph{time-bounded exclusion} $\to$
        \texttt{exclusionIsTimeBounded}).
\end{enumerate}
A keyphrase is considered covered if \emph{any} method produces a match.

\begin{figure}
    \centering
    \includegraphics[width=1\linewidth]{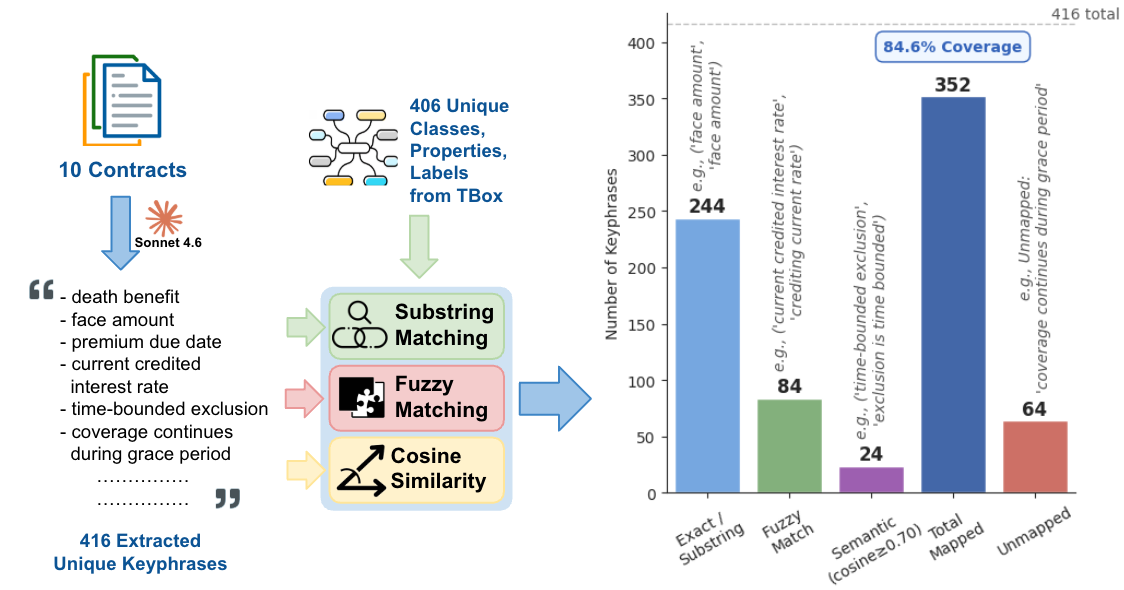}
    \caption{\textbf{Ontology Coverage Analysis.} Extracted keyphrases from the contracts are matched against the ontology TBox artifacts (classes, properties, labels) using three different matching techniques. Example tuples $(s, o)$ denote the \textbf{source contract term} ($s$) and its corresponding \textbf{mapped ontology artifact} ($o$).}
    \label{fig:coverage}
\end{figure}

As shown in Figure~\ref{fig:coverage}, 352 of 416 keyphrases
(84.6\%) were successfully mapped.
The majority were captured by exact or substring matching (244),
with fuzzy matching and semantic similarity contributing a further
84 and 24 matches respectively. 

However, manual inspection shows that the 64 keyphrases that were not matched by any automated method do not necessarily represent gaps in the ontology. We found that in most of these cases, the concept is captured
by the TBox under a different label or through a richer representational
structure that the string-level matching pipeline cannot resolve.
A concrete example is the keyphrase \emph{`coverage continues during grace
period'}, which no method maps to an ontology label. Yet the \texttt{rdfs:comment} of class \texttt{PremiumNonpaymentGracePeriod}
explicitly states:
\begin{quote}\small\itshape
  ``Triggered by failure to pay a scheduled premium.
  \textbf{Coverage continues} for specified days after premium due date.
  C1: 31 days, C2: 31 days, C5: 30 days, C7: 31 days \ldots, C9: 31 days.''
\end{quote}
The semantic content of the keyphrase is fully encoded; it is the
surface-form divergence between the phrase and the class label that
causes the miss.
Similar patterns hold for other flagged terms: \emph{lump sum payment}
is captured by \texttt{SettlementOptionProvision} with a
\texttt{coverageSourceText} annotation stating the default lump-sum
payout; \emph{proof of death} maps to \texttt{DeathCertificate} under
the \texttt{ClaimRequirement} hierarchy; \emph{outstanding loan} is
encoded through \texttt{PolicyLoan} and the \texttt{deathBenefitFormula}
property.
The coverage analysis therefore confirms that the ontology is
comprehensively aligned with the contract corpus.

\subsection{Competency Question Suite}
To evaluate the task-readiness of the TBox and ABox, we developed a suite of \textbf{58 structured scenarios} that represent common competency questions a stakeholder might ask when performing gap and overlap analysis. These scenarios are organized into the following key categories: Basic Death Claims, Loans and Conversions, Exclusions and Riders, Joint/Special Features, Advanced Product Mechanics, and Expert Provided Edge Cases. 

Each scenario follows a standardized JSON structure designed for systematic benchmarking. Each entry includes:
\begin{itemize}
    \item \textbf{Scenario Description}: A natural language prompt detailing a specific insurance event (e.g., SCEN-003: ``Insured dies by suicide exactly 13 months after the policy issue date. All premiums paid on time.'').
    \item \textbf{Double-Query Ground Truth}: Two distinct SPARQL queries, one to identify contracts where the claim is \textbf{covered} and one where it is \textbf{denied}.
    \item \textbf{Contract-Level Outcomes}: Manually verified ground truth for each of the ten contracts (C1--C10), detailing the specific status: COVERED, DENIED, or NOT\_APPLICABLE.
    \item \textbf{Traceability Evidence}: For every outcome, we include the verbatim \texttt{sourceText} and \texttt{sourceSection} from the original contract to ensure results are auditable and explainable.
\end{itemize}

By providing two SPARQL queries per scenario, we enable a robust deterministic evaluation that measures the accuracy of both positive and negative claim determinations with evidence-based traceability. 

The suite was initially generated using the same LLM-assisted workflow as the contracts, followed by \textbf{manual inspection} to ensure they are realistic and cover a good portion of the cases within the contracts. To further ensure domain accuracy, we had the suite \textbf{reviewed by a domain expert} and included several specialized test cases (8 out of the 58 total). These additions probe complex legal and administrative nuances, such as simultaneous death determination, lapses during grace periods on joint policies, the impact of pre-existing conditions on accelerated benefits.

\section{Analysis Through Use Case: Gap and Overlap Analysis}

This section operationalizes the benchmark through a concrete downstream use case: \emph{gap and overlap analysis} across a set of insurance contracts. The goal is not to prescribe a single \textit{correct} notion of gap/overlap, but to show that once contracts are represented using an ontology (TBox) with populated instances (ABox) and paired with executable scenario queries, the dataset supports precise, repeatable such analyses.

In this paper, \emph{gap} and \emph{overlap} are defined \emph{relative to a scenario} $s$ (e.g., a claim situation with conditions such as timing, cause, and policy state) and a contract set $\mathcal{C}$. The scenarios may originate from different perspectives.

From a policyholder perspective, an insured typically asks: ``If this scenario happens to me, which contracts would help me, and where would I be unprotected?''
We define:
\begin{equation*}
\mathrm{Overlap}_{\text{insured}}(s) \;=\; \{\, c \in \mathcal{C} \mid s \text{ is \textsc{Covered} under } c \,\},
\end{equation*}
i.e., \emph{overlap} is the set of contracts that cover the scenario. Intuitively, overlap indicates multiple alternatives (or redundancies) that would pay out under the same scenario.

We define the insured-centric \emph{gap} as the set of contracts that do \emph{not} provide protection for the scenario, either due to explicit denial or because the scenario does not meaningfully apply under that product's structure:
\begin{equation*}
\mathrm{Gap}_{\text{insured}}(s) \;=\; \{\, c \in \mathcal{C} \mid s \text{ is \textsc{Denied} or \textsc{Not\_Applicable} under } c \,\}.
\end{equation*}
This definition is scenario-centric: \textit{gap} is not \textit{missing text}, but rather \emph{lack of coverage for the insured under the scenario}, which can arise via exclusionary conditions or structural non-applicability.

Other stakeholders such as analysts, product designers, or regulators might ask: ``How consistently are certain risks addressed across products, and where do products diverge?'' In this view, gap/overlap can be defined in multiple ways, for example: overlap as common coverage across many contracts, overlap as common denial patterns (shared exclusion regimes), or gap as portfolio absence (no contract covers). 

However, our scenario set is primarily insured-centric. Since the benchmark releases the full set of aligned artifacts: contract text, a shared TBox, contract-specific ABox populations, and executable scenario queries with clause-level annotation, additional viewpoints can be incorporated by simply adding new scenarios and corresponding query pairs. In this sense, the benchmark does not claim to exhaust all perspectives in its current scenario set; rather, it provides a representation and evaluation substrate that makes such extensions straightforward and interoperable with the existing resources.

\subsection{Query Outcomes}

For each scenario $s$, the benchmark provides two SPARQL queries:
(i) a coverage query $Q^{+}_{s}$ that retrieves contracts for which the scenario is \textsc{Covered}, and
(ii) a denial query $Q^{-}_{s}$ that retrieves contracts for which the scenario is \textsc{Denied}. Thus, for each scenario–contract pair  $(s, c)$, the benchmark assigns exactly one outcome label: \textsc{COVERED}, \textsc{DENIED}, or \textsc{NOT\_APPLICABLE}. Collecting these labels over all scenarios and contracts yields a scenario-by-contract outcome matrix ($M$). Let $\mathrm{Ans}(Q)$ denote the set of contracts returned by query $Q$. The outcome for each contract $c \in \mathcal{C}$ is computed as:
\begin{align*}
M[s,c] &= \textsc{Covered} \quad &&\text{if } c \in \mathrm{Ans}(Q^{+}_{s}),\\
M[s,c] &= \textsc{Denied} \quad &&\text{if } c \in \mathrm{Ans}(Q^{-}_{s}),\\
M[s,c] &= \textsc{Not\_Applicable} \quad &&\text{if } c \notin \mathrm{Ans}(Q^{+}_{s}) \;\wedge\; c \notin \mathrm{Ans}(Q^{-}_{s}).
\end{align*}

In this design, \textsc{Not\_Applicable} is not a ``third query,'' but the residual class induced by the pair $(Q^{+}_{s}, Q^{-}_{s})$. It captures situations where the scenario does not match the product structure or the relevant mechanism is outside the contract's modeled scope for that scenario family.

\subsection{Gap/Overlap Summaries}

Executing the scenario queries over all $s \in \mathcal{S}$ yields a scenario-by-contract outcome matrix:
\begin{equation*}
M[s,c] \in \{\textsc{Covered}, \textsc{Denied}, \textsc{Not\_Applicable}\}.
\end{equation*}
From $M$, insured-centric overlap and gap can be computed directly:
\begin{align*}
|\mathrm{Overlap}_{\text{insured}}(s)| &= |\{\, c \in \mathcal{C} \mid M[s,c]=\textsc{Covered}\,\}|,\\
|\mathrm{Gap}_{\text{insured}}(s)| &= |\{\, c \in \mathcal{C} \mid M[s,c]\in\{\textsc{Denied},\textsc{Not\_Applicable}\}\,\}|.
\end{align*}

At the same time, the same matrix supports other stakeholder summaries without modifying the benchmark, e.g., \emph{denial overlap} (contracts that share denial outcomes for $s$), \emph{portfolio coverage gaps} (no contract covers a scenario),
\begin{equation*}
\forall c \in \mathcal{C},\; M[s,c]\neq \textsc{Covered},
\end{equation*}
or patterns of structural non-applicability (scenarios systematically \textsc{Not\_Applicable} for certain product families), reflecting genuine product design differences rather than errors. The key point is that the benchmark provides a foundation (explicit structure plus executable scenario queries) upon which multiple gap/overlap interpretations can be layered.

\subsection{Comparison with a Text-Only LLM Pipeline}

To assess how well current large language models perform on the task, we evaluated the same 58 scenarios using a text-only LLM pipeline. We used three state-of-the-art LLMs: \textit{ChatGPT-5.3}~\cite{openai2026chatgpt53}, \textit{Gemini-3}~\cite{google2025gemini3}, and \textit{Claude Sonnet-4.6}~\cite{anthropic2026claude46}. Each model was accessed through its public web interface and prompted with the same structured instructions to analyze the contracts and scenarios. The prompt required the model to determine, for every contract-scenario pair, whether the scenario was \texttt{COVERED}, \texttt{DENIED}, or \texttt{NOT\_APPLICABLE}, and to support the classification with a relevant clause or excerpt from the contract text.

The 58 scenarios, applied to 10 contracts, produces \textbf{580 contract-scenario evaluations per model}. Since three models were tested, the evaluation comprises \textbf{1,740 total classifications}. Model outputs were parsed into the required JSON format and compared against the ground-truth labels derived from the ontology-backed contract representations.

Table~\ref{tab:llm-overall-accuracy} summarizes the overall accuracy of each model. Claude achieved the strongest performance at \textbf{87.76\%} ($509/580$), followed by ChatGPT at \textbf{72.93\%} ($423/580$) and Gemini at \textbf{65.17\%} ($378/580$). While these results indicate that these LLMs can mostly understand contracts terms, a more detailed analysis reveals variability in both reasoning behavior and interpretation of contractual scope.

\begin{table}[b]
\centering
\caption{Overall accuracy of the evaluated LLMs on the benchmark.}
\label{tab:llm-overall-accuracy}
\begin{tabular}{lc}
\toprule
\textbf{Model} & \textbf{Accuracy} \\
\midrule
Claude Sonnet 4.6 & \textbf{87.76\%} ($509/580$) \\
ChatGPT-5.3 & 72.93\% ($423/580$) \\
Gemini-3 & 65.17\% ($378/580$) \\
\bottomrule
\end{tabular}
\end{table}

\begin{wrapfigure}{r}{0.65\textwidth}
    \vspace{-4ex}
    \centering
    \includegraphics[width=1\linewidth]{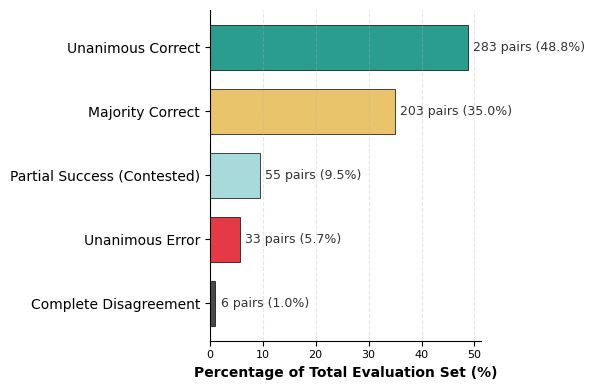}
    \caption{Inter-LLM agreement patterns across the 580 contract-scenario pairs.}
    \label{fig:inter_llm_agreement}
\end{wrapfigure}

A useful way to further understand this variability is through \textbf{inter-model agreement}, which captures whether the models arrive at the same judgment for a given contract--scenario pair. As shown in Figure~\ref{fig:inter_llm_agreement}, among the 580 evaluated pairs, the models produced identical correct answers in \textbf{283 cases (48.8\%)}, while \textbf{203 cases (35.0\%)} were resolved correctly by a majority of models but not unanimously. At the same time, \textbf{55 cases (9.5\%)} produced mixed outcomes, \textbf{33 cases (5.7\%)} resulted in unanimous but incorrect answers, and \textbf{6 cases (1.0\%)} exhibited complete disagreement across all three models. These results indicate that while LLMs often agree on straightforward scenarios, their reasoning becomes less stable when scenarios are complex.

Further insight emerges when examining which contracts produce the largest number of errors. For Claude, the five most error-prone contracts are C8, C9, C6, C10, and C4. For ChatGPT, they are C8, C4, C10, C7, and C9. For Gemini, they are C6, C1, C9, C8, and C7. Notably, \textbf{C4} (Variable Universal Life), \textbf{C8} (Joint Survivorship), and \textbf{C10} (Indexed Universal Life), \textbf{the three most complex contracts} in the benchmark \textbf{all appear among the top-five most error-prone contracts} for a majority of the evaluated models. These results suggest, as contractual complexity increases, LLMs become more prone to misinterpretation or incomplete reasoning.

To better understand the nature of these failures, we examined the distribution of mismatch types between model predictions and ground truth. 

\begin{table}[t]
\centering
\caption{Most frequent mismatch types between ground truth and LLM predictions.}
\label{tab:llm-mismatches}
\begin{tabular}{lc}
\toprule
\textbf{Mismatch type} (\texttt{TRUE} $\rightarrow$ \texttt{PRED.}) & \textbf{Count} \\
\midrule
\texttt{NOT\_APPLICABLE} $\rightarrow$ \texttt{DENIED} & 207 \\
\texttt{NOT\_APPLICABLE} $\rightarrow$ \texttt{COVERED} & 126 \\
\texttt{COVERED} $\rightarrow$ \texttt{DENIED} & 30 \\
\texttt{COVERED} $\rightarrow$ \texttt{NOT\_APPLICABLE} & 30 \\
\texttt{DENIED} $\rightarrow$ \texttt{COVERED} & 23 \\
\bottomrule
\end{tabular}
\end{table}

As shown in Table~\ref{tab:llm-mismatches}, the most common errors are \texttt{NOT\_APPLICABLE} $\rightarrow$ \texttt{DENIED} (\textbf{207}) and \texttt{NOT\_APPLICABLE} $\rightarrow$ \texttt{COVERED} (\textbf{126}), followed by much smaller numbers of \texttt{COVERED} $\rightarrow$ \texttt{DENIED} (\textbf{30}), \texttt{COVERED} $\rightarrow$ \texttt{NOT\_APPLICABLE} (\textbf{30}), and \texttt{DENIED} $\rightarrow$ \texttt{COVERED} (\textbf{23}). The dominance of the \texttt{NOT\_APPLICABLE} $\rightarrow$ \texttt{DENIED} pattern highlights a recurring difference between the benchmark's contract interpretation logic and the reasoning heuristics used by LLMs. In the benchmark annotations, \texttt{NOT\_APPLICABLE} indicates that the contract does not contain clauses governing the scenario. In contrast, LLMs frequently interpret the absence of such clauses as grounds for denying the claim. 

For example, for the scenario: \emph{``The insured dies in a car accident. Toxicology confirms the insured had a blood alcohol concentration above the legal limit at the time of the accident. The beneficiary files an accidental death benefit claim''}, as the ground truth, most contracts classify this scenario as \texttt{COVERED}. The reason is that these policies do \emph{not} include an accidental death rider or a drug/alcohol \textbf{exclusion} clause, meaning that the standard death benefit remains payable. However, many LLM responses classified the scenario as \texttt{DENIED} or \texttt{NOT\_APPLICABLE}, apparently because of no mention of intoxication-related exclusions. This also helps explain the observed \texttt{COVERED} $\rightarrow$ \texttt{NOT\_APPLICABLE} and \texttt{COVERED} $\rightarrow$ \texttt{DENIED} mismatches.

A second example illustrates how inter-LLM disagreement becomes especially visible when the scenario depends on a specific policy structure. Consider the scenario: \emph{``On a joint survivorship policy, the first insured dies with an outstanding policy loan of \$50,000 at 5.5\% annual interest. The surviving insured and trustee want to know how the loan is treated during the continuation period and its impact on the eventual death benefit.''} This is \textbf{one of the benchmark's eight special-case} scenarios. Only \textbf{Contract C8} is a \emph{joint survivorship policy}, so the correct labeling is \texttt{COVERED} for C8 and \texttt{NOT\_APPLICABLE} for all other contracts. In this case, the three models exhibited completely different reasoning behaviors. Claude correctly recognized the structural constraint, identified that the scenario is meaningful only for the joint policy, and cited the relevant loan clauses governing treatment before and after the first death. ChatGPT instead marked nearly all contracts as \texttt{COVERED}, apparently inferring that any death-benefit clause was sufficient for applicability, while Gemini marked most contracts as \texttt{DENIED}, often citing generic product descriptions that do not substantively justify the classification. This example demonstrates a setting in which the models diverge completely in how they interpret applicability, coverage, and supporting evidence.

A third representative case explains why \texttt{NOT\_APPLICABLE} $\rightarrow$ \texttt{DENIED} is by far the most common mismatch. Consider the scenario: \emph{``The policy owner wishes to take out a loan against the policy's cash value or account value shortly after the policy has been in force for one year.''} For Contract \textbf{C9}, the correct label is \texttt{NOT\_APPLICABLE}, because the contract does not have any clauses regarding policy loans. However, all three LLMs classified the scenario as \texttt{DENIED}, using the absence of such clauses as justification. This suggests that the models are not necessarily missing the relevant text; rather, they are applying a different decision logic from the benchmark. 

Beyond the classification errors themselves, the examples above reveal another important pattern: the \textbf{quality of evidence citations} provided by LLMs varies substantially. In several cases, the cited text is only loosely connected to the assigned label, such as generic product descriptions or broad death-benefit statements. This is especially visible in the joint survivorship example, where Gemini and ChatGPT often cited short descriptions like ``term life insurance'' or ``whole life insurance'' for contracts to which the scenario does not even apply. Such outputs may appear superficially well-formed, but they are not evidence-grounded in the sense required for reproducible analysis.

Taken together, these findings highlight a fundamental distinction between \textbf{ontology-driven reasoning} and \textbf{LLM-based interpretation}. In the ontology-based approach described earlier, scenarios are evaluated through explicit queries over formally defined policy concepts and relationships. Because these representations encode the structure of the contract, query results depend solely on the modeled terms present in the policy. In contrast, LLMs may incorporate implicit assumptions, background knowledge, or heuristic interpretations that lead to plausible but unsupported conclusions. This distinction helps explain why LLMs can perform reasonably well on straightforward cases while remaining inconsistent on cases that require strict handling of contractual scope and structural constraints.

\subsection{Gap/Overlap Analysis as a Test of KG Task Readiness}

The analysis above speaks directly to the claim made in the introduction: that
gap and overlap analysis serves as a useful test of \emph{task readiness} for
a KG, asking whether an ontology-based representation can surface
coverage differences consistently and supply evidence for each outcome. The
SPARQL-driven pipeline answers both parts of that question affirmatively and in
a way that the text-only LLM baseline cannot match. Consistency is guaranteed
structurally: because the ABox encodes each relevant provision as a typed,
property-bearing individual with explicit domain and range constraints, the same
query logic applies uniformly across all ten contracts, with no variation in how
parametric differences such as exclusion periods, grace period triggers, benefit
outcome types, etc., are evaluated. Evidence is produced automatically and
traceably: every query result carries the \texttt{sourceContractID},
\texttt{sourceContractSection}, and \texttt{coverageSourceText} of the
supporting clause, so each outcome is not merely asserted but justified by the
originating contract text. We note that this consistency is a direct consequence of the manual knowledge modeling effort, and that is exactly what we require for a reproducible and auditable benchmark.

As demonstrated in the previous section, the LLM pipeline, operating on unstructured text, can approximate coverage determinations in straightforward cases but cannot guarantee consistency for complex cases, and cannot always produce structured evidence trails. The difference between the two approaches is therefore not merely a gap in language understanding, but a gap in \emph{representational commitment}: the ontology forces the modeler to resolve ambiguities, enumerate
value-partition members, and declare domain and range constraints at construction
time. It is that modeling effort that enables deterministic, evidence-linked analysis at query time. In this sense, the benchmark validates the original motivation: structured KG representation does not merely organize contract knowledge, it makes that knowledge queriable with evidence.

\section{Conclusion and Future Work}

This paper presents an executable benchmark for KG quality evaluation through
gap and overlap analysis on life insurance contracts. The benchmark aligns three
resources: ten domain expert verified synthetic contracts, a domain ontology
(TBox) with a fully populated and evidence-annotated ABox in which every
instance is anchored to a contract identifier, section reference, and verbatim
clause text, and 58 structured scenarios with SPARQL-based ground truth and
clause-level annotation. We demonstrated the benchmark through a gap and overlap
use case, showing that the ontology-driven pipeline produces deterministic,
fully traceable coverage determinations across all ten contracts, and compared
it against a text-only LLM baseline to illustrate where unstructured inference
degrades and why representational commitment matters for consistent, auditable
analysis.

However, we acknowledge several limitations of the present benchmark. The contracts are synthetic, although expert-reviewed for legal plausibility and internal consistency, external validity may be limited when compared to proprietary, naturally occurring contracts. Also, the benchmark is instantiated particularly in life insurance, though the methodology is applicable beyond this domain. The ontology is purpose-built to serve the benchmark rather than as a general domain ontology, and extending it toward broader coverage remains a natural direction for future work. Moreover, the ontology was bootstrapped with LLM assistance, which may introduce some modeling bias despite multi-model review and manual validation. Additionally, the scenario set is diverse but not exhaustive: it can be extended to broaden coverage. Finally, the LLM evaluation uses a text-only baseline with fixed prompting, and stronger prompting or retrieval-augmented methods may perform differently.

Beyond gap and overlap analysis, the benchmark's aligned resources support
a range of downstream tasks. The contract corpus paired with the verified TBox
and ABox provides a natural testbed for ontology learning and KG population
methods, where automatically constructed representations can be evaluated
against the ground truth end-to-end. The clause-level annotations
make the benchmark directly applicable to evidence-grounded question answering
and explainable retrieval tasks. The scenario suite, with its structured
outcomes across ten contracts, also supports contract reasoning and
entailment tasks in the legal text processing research. More broadly, the benchmark's
structure: aligned document text, a verified TBox, an annotated
ABox, and executable scenario queries, is a domain-agnostic template, and
extending it to other policy-like domains such as healthcare coverage or
financial regulation is a natural next step toward a shared infrastructure
for KG quality evaluation across legal and regulatory document collections. It also opens a direction toward hybrid neuro-symbolic pipelines. Our evaluation deliberately contrasts a pure neural approach with a pure symbolic one, but the two need not be mutually exclusive: an LLM could extract structured features that drive ontology-backed reasoning, or translate natural language scenarios into formal SPARQL queries, combining the language fluency of neural models with the consistency and traceability guarantees of the KG. The benchmark's clause-level ground truth and executable queries are the resources needed to develop and evaluate such approaches, and this could be a promising direction for future work.

\section*{Acknowledgments}
We acknowledge the support from NSF IUCRC CRAFT center research grant (CRAFT Grant \#22022) for this research. The opinions expressed in this publication do not necessarily represent the views of NSF IUCRC CRAFT. We are also grateful for the advice and resources from our CRAFT Industry Board members in shaping this work.

\bibliography{references}

\appendix

\end{document}